\documentclass[letterpaper, 10 pt, conference]{ieeeconf}

\IEEEoverridecommandlockouts
\usepackage{graphics}
\usepackage{epsfig}
\usepackage{mathptmx}
\usepackage{times}
\usepackage{amsmath}
\usepackage{amssymb}
\usepackage{algorithm}
\usepackage{algorithmic}
\usepackage{subcaption}
\usepackage{xurl}
\usepackage{hyperref}
\usepackage{bm}

\title{\LARGE \bf
Legible Shared Autonomy: Implicit Communication of Robot Belief through Motion
}

\author{Jinwei Liu$^{1}$, Pengfei Li$^{1}$, Shaofeng Chen$^{2}$, Tao Wang$^{2,3}$ and Yun-Bo Zhao$^{1}$
\thanks{This work was supported by the National Natural Science Foundation of China under Grant 62473352, Grant 62403445, and Grant 62503147; 
the State Key Laboratory of Autonomous Intelligent Unmanned Systems. The opening project number is ZZKF2025-1-10;
the Anhui Provincial Natural Science Foundation under Grant 2508085QF230.}
\thanks{(Corresponding author: Pengfei Li)}
\thanks{$^{1}$Jinwei Liu and Pengfei Li are with the Department of Automation, University of Science and Technology of China, Hefei 230027, China
{\tt\small liujinwei@mail.ustc.edu.cn}
{\tt\small puffylee@ustc.edu.cn}}
\thanks{$^{2}$Shaofeng Chen is with the Department of Automation, Hefei University of Technology, Hefei 230009, China
{\tt\small chenshaofeng@hfut.edu.cn}}
\thanks{$^{2,3}$Tao Wang is with the Department of Automation, Hefei University of Technology, Hefei 230009, China, and also with the National Key Laboratory of Autonomous Intelligent Unmanned Systems, Beijing Institute of Technology, Beijing 100081, China
{\tt\small twang@hfut.edu.cn}}
\thanks{$^{1}$Yun-Bo Zhao is with the Department of Automation, University of Science and Technology of China, Hefei 230027, China
{\tt\small ybzhao@ustc.edu.cn}}
}
\begin{document}

\maketitle
\thispagestyle{empty}
\pagestyle{empty}

\begin{abstract}
Shared autonomy systems combine user input with autonomous assistance to help users with motor impairments control robot arms to perform everyday manipulation tasks, by inferring user goals and providing appropriate guidance. However, the robot's internal beliefs about user goals cannot be observed by users. Traditional shared autonomy systems provide assistance along efficient shortest paths toward inferred goals, but when multiple objects lie in similar directions, such assistive motion remains ambiguous and fails to reveal the specific goal identified by the robot. This creates two critical problems. First, when the robot correctly infers the goal, users continue controlling because they cannot perceive understanding from ambiguous assistive motion, wasting effort when autonomous completion would suffice. Second, when the robot misunderstands intent, users cannot quickly detect errors until assistive motion diverges significantly, requiring substantial corrective input. We address this by introducing legible motion into shared autonomy, where robot actions must both advance toward the goal and clearly reveal which goal has been inferred, enabling users to understand the robot's beliefs and adjust control accordingly. The robot modulates communication strength through confidence-aware adaptive authority allocation by providing assertive legible assistive actions when confident while increasing user authority when uncertain, transforming shared autonomy into transparent bidirectional collaboration. User studies including simulation, and physical experiments with a six-degree-of-freedom robot arm demonstrate that legible shared autonomy significantly improves users' understanding of robot beliefs and reduces user control effort compared to standard shared autonomy.
\end{abstract}

\section{INTRODUCTION}

Shared autonomy has emerged as a promising approach for enabling individuals with motor impairments to perform everyday manipulation tasks such as grasping objects and self-feeding with assistive robot arms\cite{javdani2015shared, gopinath2016human, nanavati2023physically, collier2025sense, losey2018review, di2025semi}. These systems combine user input with autonomous assistance, where users provide commands through input devices while the robot infers their intended goals and provides assistance. This collaboration reduces the control burden on users while preserving their sense of agency\cite{collier2025sense}. Despite these advantages, current shared autonomy systems face a fundamental transparency problem that limits their effectiveness.

\begin{figure}
    \centering
    \includegraphics[width=1\linewidth]{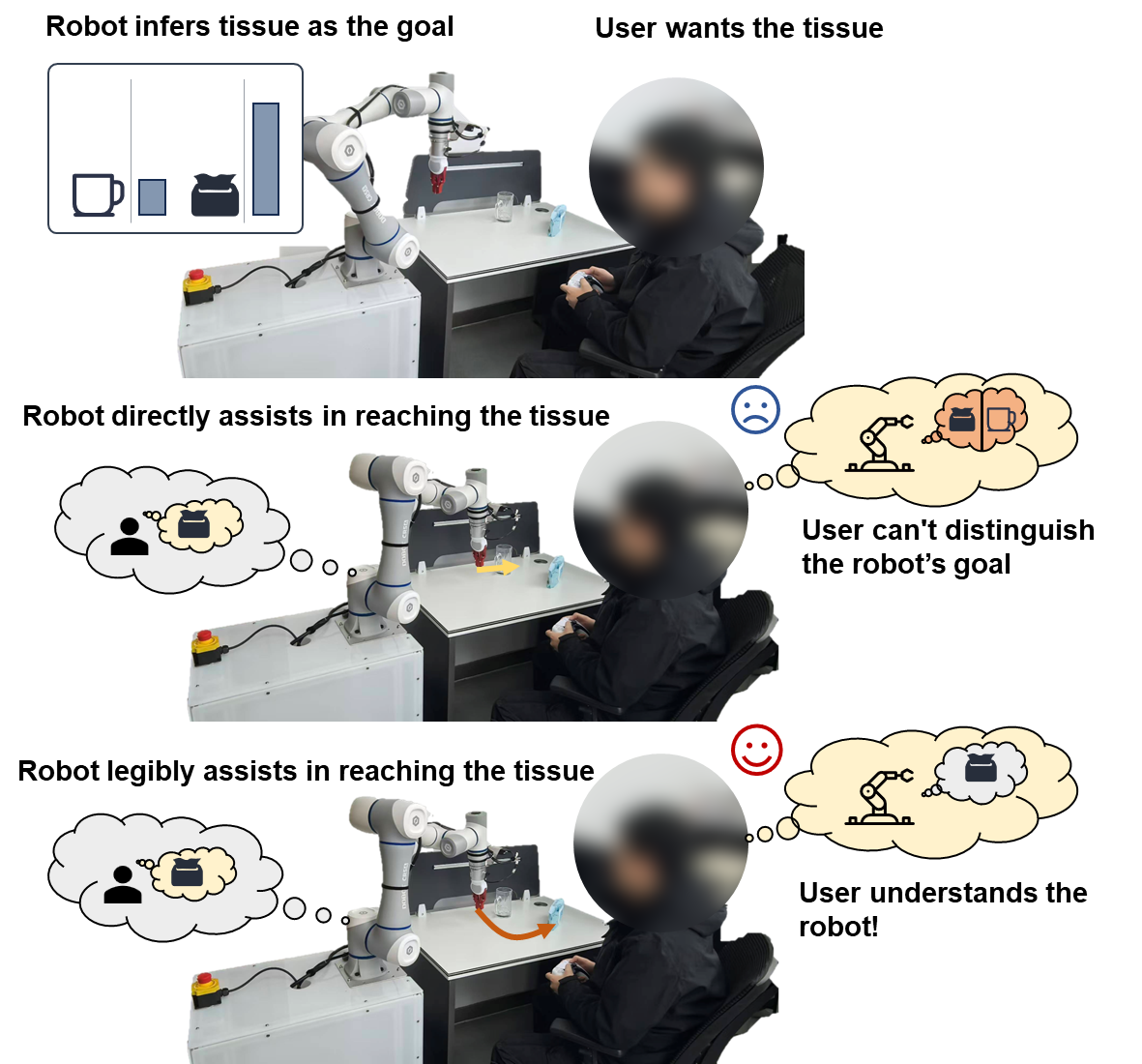}
    \caption{(Top) The robot infers that the user wants the tissue. (Middle) The robot starts assisting in reaching the tissue, but the user is still uncertain and continues to control the joystick. (Bottom) The robot moves clearly toward the tissue, showing its legible motion, which allows the user to understand the robot's goal inference and adjust their control accordingly.}
    \label{fig:disc}
\end{figure}

Consider a user reaching for a tissue on a cluttered table with a cup nearby, as illustrated in Figure~\ref{fig:disc}. The robot maintains an internal belief about the user's goal based on observed inputs\cite{jain2019probabilistic} and infers with high confidence that the goal is the tissue, but the user has no way to know this. Traditional shared autonomy\cite{javdani2015shared, dragan2013policy, guleccyuz2025enhancing} would provide assistance in the most efficient manner toward the tissue, yet this same direction could also lead toward the cup. Should the user continue controlling or release the joystick? Without access to the robot's belief, they face a fundamental dilemma: they cannot distinguish whether the robot's assistance aligns with their intent or not, forcing them to maintain vigilant control throughout the task.

Specifically, this lack of transparency manifests in two critical failure modes. First, when the robot correctly infers the user's goal, users continue providing control input because they cannot perceive that the robot has understood their intent, resulting in unnecessary operational burden even when the robot could autonomously complete the task. Second, when the robot misunderstands the user's intent, users cannot quickly detect this error until the system's motion diverges significantly from their desired trajectory, requiring substantial corrective input.

Our key insight is that robots can communicate their understanding of user intent through the character of their assistive motion. Rather than moving efficiently but ambiguously, robots can select actions that clearly discriminate the inferred goal from alternatives, moving in a manner that unmistakably indicates tissue versus cup when those are the options. This approach adapts legible motion\cite{dragan2013legibility, bronars2024legibility} concepts from communicating the robot's own goals to revealing the robot's beliefs about the user's goals. Our approach balances task assistance with motion clarity by simultaneously optimizing two objectives: how effectively an action advances toward the inferred goal, and how unambiguously it reveals which goal has been inferred. By observing the robot's motion, users can dynamically adjust their control strategy: reducing input when the motion confirms correct understanding, or immediately providing corrections when it reveals a misunderstanding. This motion-based communication transforms shared autonomy from one-way inference into bidirectional collaboration, closing the transparency gap without requiring explicit feedback interfaces.

Our contributions are as follows:

First, we propose an action-level legibility metric suitable for real-time assistance. Unlike prior trajectory-level approaches that optimize complete paths offline, our metric can be computed efficiently at each timestep, enabling robots to achieve legible intent communication in interactive settings where they must respond immediately to unpredictable user inputs.

Second, we introduce legibility into the shared autonomy framework for the first time, enabling robots to reveal their inferred beliefs about user goals through assistive motion. Additionally, we design an authority allocation strategy that allows the robot to provide clear, assertive assistance when highly confident while increasing user control authority when uncertain, achieving transparent collaboration without compromising task performance.

Third, we validate our approach through user studies including 2D simulation with 20 participants and physical experiments using a six-degree-of-freedom robot arm with 15 participants, demonstrating that legible shared autonomy significantly improves users' understanding of robot beliefs and reduces user control effort compared to standard shared autonomy.

\section{RELATED WORK}
\label{sec:related}
The following sections review related work in legible motion planning and shared autonomy systems.

\subsection{Legible Motion Planning}
Legibility in robot motion was formalized by Dragan et al.~\cite{dragan2013legibility} as trajectory optimization to maximize the probability of observers correctly inferring the robot's goal. Early approaches relied on manual cost function design, while recent work has automated legibility generation through learning-based frameworks. Inverse reinforcement learning~\cite{zeng2025automate} and reinforcement learning~\cite{busch2017learning} extract legibility features directly from data, addressing the complexity of manual tuning. Generative models have further enriched expressiveness: diffusion-based frameworks~\cite{shi2025controlling, bronars2024legibility} enable continuous modulation from ambiguous to distinct motion, generating trajectories with varying degrees of legibility rather than a single optimal solution.

However, a fundamental distinction exists between prior work and ours. Existing methods address legibility in general human-robot collaboration, where the robot communicates its own intent ($G_{robot} \to \text{User}$). In shared autonomy, the robot lacks its own goals and must instead reveal its beliefs about user intent ($b(G_{user}) \to \text{User}$), encompassing both goal identity and inference confidence. We bridge this gap by shifting from legible intent to legible belief, where action-level legibility reveals the inferred goal, while adaptive authority allocation signals confidence.

\subsection{Shared Autonomy}
Shared autonomy systems enable users with motor impairments to control robotic manipulators through the combination of user input and autonomous assistance\cite{javdani2015shared, gopinath2016human, nanavati2023physically, collier2025sense, losey2018review}. The fundamental mechanism involves inferring user intent from observed inputs and providing assistance toward the most probable goal. Early approaches focused on fixed blending policies\cite{levine1999navchair} and arbitration frameworks\cite{dragan2013policy}, while recent work has integrated probabilistic intent recognition\cite{jain2019probabilistic} with adaptive autonomy allocation\cite{broad2016trust} to dynamically modulate assistance strength. Contemporary advances in deep reinforcement learning\cite{reddy2018shared} and diffusion models\cite{yoneda2023diffusha} have enabled end-to-end learned policies that substantially improve assistance quality and smoothness.

Despite these advances, existing methods share a fundamental limitation: they prioritize task performance optimization at the expense of system transparency. Users cannot observe how their inputs are interpreted or understand the robot's internal belief state, creating a transparency gap where effective assistance comes at the cost of comprehensibility. Prior work has attempted to address this gap through explicit feedback channels, including visual interfaces\cite{hoegerman2024aligning}. While these approaches effectively convey information, they require additional hardware and, more critically, demand split attention between control execution and feedback monitoring.

This work proposes motion-based implicit communication as an integrated alternative. Rather than relying on auxiliary interfaces, the robot's assistive motion itself serves as the communication channel. Robots can reveal their inference state through motion characteristics, which enables users to directly perceive both goal predictions and confidence levels from observed assistance, minimizing cognitive overhead while maintaining transparency.

\section{METHODOLOGY}
\label{sec:method}

We introduce a legible shared autonomy framework in which robots reveal their beliefs about user intent through the characteristics of assistive motion. This section presents our methodology. First, we formalize the shared autonomy problem and define the goal-oriented belief update process. Subsequently, we introduce a user action model that captures goal-directed behavior and define action-level legibility based on this model. Building on this foundation, we formulate robot action selection as a dual optimization problem that balances legibility and task performance. Finally, we describe an adaptive authority allocation strategy that combines belief confidence.

\subsection{Problem Formulation}
\label{sec:problem}

The shared autonomy setting involves a user and robot jointly controlling a dynamical system. At each timestep $t$, the user provides action $\bm{a}_H^t \in \mathcal{A}_H$ through an input device specifying desired velocity or direction. The robot generates assistive action $\bm{a}_R^t \in \mathcal{A}_R$ based on inferred user goals. System state $\bm{s}^t \in \mathcal{S}$ evolves according to standard dynamics driven by the combined action $\bm{a}^t$:
\begin{equation}
\bm{s}^{t+1} = f(\bm{s}^t, \bm{a}^t), \quad 
\bm{a}^t = g(\bm{a}_H^t, \bm{a}_R^t)
\label{eq:system_dynamics}
\end{equation}
where $g(\cdot)$ is the authority allocation function that combines user and robot actions, defined in Section~\ref{sec:adaptive_authority}.

To provide effective assistance, the robot must infer what the user is trying to accomplish. However, the robot does not directly observe the user's intended goal $\theta \in \Theta$, where $\Theta$ represents the discrete set of candidate goals such as objects the user might reach for. Instead, the robot maintains a belief distribution $b^t(\theta)$ over possible goals, which it updates through Bayesian inference based on observed user actions:

\begin{equation}
b^{t+1}(\theta) \propto \pi_H(\bm{a}_H^t \mid \bm{s}^t, \theta) \cdot b^t(\theta)
\label{eq:belief_update}
\end{equation}
where $\pi_H(\bm{a}_H \mid \bm{s}, \theta)$ models how a user pursuing goal $\theta$ would act in state $\bm{s}$.

The robot then uses this belief to compute its assistive action:
\begin{equation}
\bm{a}_R^t = \pi_R(\bm{s}^t, b^t)
\label{eq:robot_action}
\end{equation}
where $\pi_R$ is the robot policy detailed in Section~\ref{sec:optimization}.

Standard shared autonomy uses this belief to silently assist toward the most likely goal, creating a communication gap. This work addresses the gap by enabling robots to communicate beliefs through legible motion, allowing users to perceive understanding and adjust control accordingly.

\subsection{User Action Model}
\label{sec:human_model}

The belief update mechanism in Eq.~\ref{eq:belief_update} relies on the user action model $\pi_H(\bm{a}_H \mid s, \theta)$, which represents how likely a user pursuing goal $\theta$ would take action $\bm{a}_H$ in state $s$. 

We employ a Boltzmann-rational model widely used in shared autonomy~\cite{jain2019probabilistic,hoegerman2024aligning}, which represents users as approximately optimal actors who balance goal achievement with effort minimization:
\begin{equation}
\pi_H(\bm{a}_H \mid s, \theta) = \frac{\exp\left(\beta_r R(\bm{a}_H, s, \theta)\right)}{Z(s, \theta)}
\label{eq:boltzmann}
\end{equation}
where $\beta_r > 0$ controls action optimality and $Z(s, \theta) = \sum_{\bm{a}'_H} \exp(\beta_r R(\bm{a}'_H, s, \theta))$ normalizes the distribution. This model provides a probability distribution over actions that a user would take in state $s$ when pursuing goal $\theta$.

The reward function captures two factors influencing user action selection, namely progress toward goals and control effort:
\begin{equation}
R(\bm{a}_H, s, \theta) = \left(\|\theta - s\| - \|\theta - (s + \Delta t \cdot \bm{a}_H)\|\right) - \lambda_e \|\bm{a}_H\|
\label{eq:reward}
\end{equation}
The first term measures distance reduction to goal $\theta$, positive when actions bring the system closer. The second term penalizes control effort, reflecting natural user preferences to minimize exertion while making progress. This formulation captures how users select actions balancing goal proximity improvement against physical effort costs. Parameter values can alternatively be learned from demonstration data when available.

\subsection{Action-Level Legibility}
\label{sec:legibility}

Prior trajectory-level approaches optimize complete paths offline by integrating goal probabilities over time, which makes them incompatible with interactive settings requiring immediate responses to unpredictable user inputs. To address this limitation, we propose an action-level legibility metric that enables efficient computation at each timestep to guide real-time action selection.

A robot action $\bm{a}_R$ demonstrates legibility with respect to an inferred goal when it clearly indicates that goal through alignment with characteristic user goal-directed behavior patterns. This discriminability is quantified by comparing the likelihood of the robot action under the inferred goal versus the most confusable alternative using the user action model. Intuitively, a legible robot action should appear much more probable when a user were pursuing the inferred goal than when pursuing any competing goal. We express this as a probability ratio:
\begin{equation}
\frac{\pi_H(\bm{a}_R \mid s^t, \theta^*)}{\max_{\theta \neq \theta^*} \pi_H(\bm{a}_R \mid s^t, \theta)}
\label{eq:legibility_ratio}
\end{equation}
where the numerator represents how likely the robot action $\bm{a}_R$ would be if a user were pursuing the inferred goal $\theta^* = \arg\max_\theta b^t(\theta)$, and the denominator represents the highest likelihood under any competing goal. A large ratio indicates the action strongly discriminates the inferred goal from alternatives.

For numerical stability and computational convenience, we take the logarithm to convert this ratio into a difference, defining our legibility metric as:
\begin{equation}
\mathcal{L}(\bm{a}_R \mid s^t, \theta^*) = \log \pi_H(\bm{a}_R \mid s^t, \theta^*) - \log \max_{\theta \neq \theta^*} \pi_H(\bm{a}_R \mid s^t, \theta)
\label{eq:legibility}
\end{equation}
This formulation provides numerical stability when probabilities are small. Positive values indicate the action is more probable under the inferred goal than any alternative, with larger values signifying stronger discriminability. Values near zero indicate the action is approximately equally likely under multiple goals, offering little discriminative information. Negative values indicate the action is actually more characteristic of a competing goal than the inferred goal, suggesting poor discriminability or potential misalignment.

\subsection{Optimization Framework}
\label{sec:optimization}

Robot action selection balances two objectives, namely communicating beliefs through legible motion and providing effective task assistance. Identifying the robot's most confident belief as $\theta^*$, the action selection formulation becomes:
\begin{equation}
\bm{a}_R^* = \arg\max_{\bm{a}_R \in \mathcal{A}_R} \left[ 
    \lambda \mathcal{L}(\bm{a}_R \mid s^t, \theta^*) + 
    Q(\bm{a}_R, s^t, \theta^*) 
\right]
\label{eq:optimization}
\end{equation}
where $\mathcal{L}$ measures action legibility, $Q$ evaluates task performance, and $\lambda \geq 0$ weights legibility importance relative to task performance. Setting $\lambda = 0$ reduces to standard shared autonomy without legible communication, while increasing $\lambda$ prioritizes legibility more strongly.

Task performance evaluation measures how close the robot's action brings the system to the goal:
\begin{equation}
Q(\bm{a}_R, s^t, \theta^*) = -\left\|\theta^* - \left(s^t + \Delta t \cdot \bm{a}_R\right)\right\|
\label{eq:task_performance}
\end{equation}
Absolute distance to goal is used rather than distance reduction because achieving proximity matters for task completion. Unlike the user reward function, robot action magnitude is not penalized since robots do not experience fatigue and can exert necessary effort to reach goals efficiently.

For continuous action spaces, discretization samples candidate directions around the task-optimal direction. The task-optimal direction is computed as the unit vector from current state toward inferred goal, with candidates sampled uniformly in a cone around this direction. Each candidate is scaled to maximum velocity magnitude, with legibility and task performance computed for each, then combined to select the optimal robot action. The computational complexity per timestep remains tractable for real-time control at typical frequencies with reasonable problem sizes.

\subsection{Adaptive Authority Allocation}
\label{sec:adaptive_authority}

To balance the control authority between the robot and the user, we dynamically adjust the authority allocation weight $\beta^t$ based on inference confidence and proximity to the goal: the robot conveys its internal beliefs with greater legibility when it has high certainty about the inferred intent, whereas user authority is retained when the robot's estimation is ambiguous. The fused control action at time step $t$ is computed by the authority allocation function: $g(\bm{a}_H^t, \bm{a}_R^t)$, as defined in the following equation:
\begin{equation}
\bm{a}^t = g(\bm{a}_H^t, \bm{a}_R^t) = \begin{cases}
\bm{a}_R^t & \text{if } \|\bm{a}_H^t\| < \epsilon \\
\beta^t \bm{a}_H^t + (1-\beta^t) \bm{a}_R^t & \text{otherwise}
\end{cases}
\label{eq:action_execution}
\end{equation}
where $\epsilon$ is a small threshold for detecting user input, and $\beta^t \in [0,1]$ is a time-varying authority allocation parameter. Larger values grant more authority to the user, while smaller values amplify the robot's legible assistance. When the user releases the input device ($\|\bm{a}_H^t\| < \epsilon$), the robot assumes full control, allowing users to delegate once they observe correct inference.

The authority allocation parameter adapts jointly to inference confidence and task progress:
\begin{equation}
\beta^t = \beta_{\text{base}} - \alpha(b^t) \cdot \gamma(d^t) \cdot (\beta_{\text{base}} - \beta_{\text{min}})
\label{eq:adaptive_beta}
\end{equation}
where $\beta_{\text{base}}$ and $\beta_{\text{min}}$ are the baseline and minimum user authority, $b^t = b^t(\theta^*)$ denotes the robot's confidence in its inferred goal $\theta^*$, and $d^t = \|\theta^* - \bm{s}^t\|$ is the current distance to that goal. The product $\alpha \cdot \gamma$ determines how far $\beta^t$ decreases from baseline toward its minimum.

The confidence modulation $\alpha$ activates robot authority only above a confidence threshold:
\begin{equation}
\alpha(b) = \max\left(0, \min\left(1, \frac{b - b_{\text{thresh}}}{1 - b_{\text{thresh}}}\right)\right)
\label{eq:confidence_mod}
\end{equation}
Below $b_{\text{thresh}}$, $\alpha = 0$ and $\beta^t$ remains at baseline, preserving full user control. Above $b_{\text{thresh}}$, $\alpha$ grows linearly toward one, progressively increasing robot authority as confidence rises.

The distance modulation $\gamma$ increases robot authority as the system approaches the goal:
\begin{equation}
\gamma(d) = \max\left(0, \min\left(1, \frac{d_{\max} - d}{d_{\max} - d_{\min}}\right)\right)
\label{eq:distance_mod}
\end{equation}
When the system is far from the goal ($d > d_{\max}$), $\gamma = 0$ and authority remains at baseline, leaving the user to guide the initial approach. As the system closes in ($d < d_{\max}$), $\gamma$ increases, reducing $\beta^t$ and granting the robot greater authority to execute precise legible motions near the goal. The thresholds $d_{\min}$ and $d_{\max}$ are set relative to workspace dimensions.

\begin{figure}[h]
    \centering
    \includegraphics[width=0.9\linewidth]{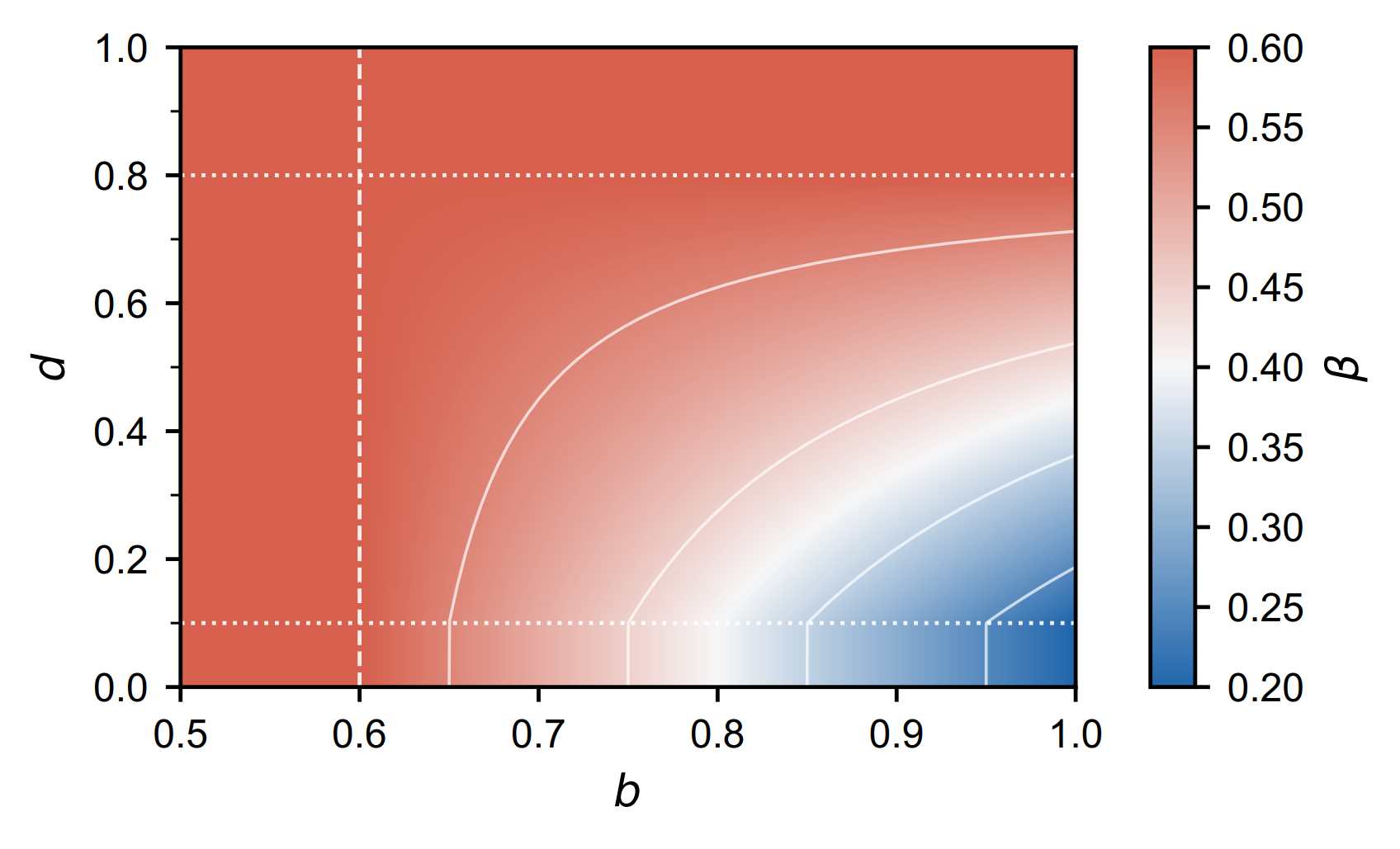}
    \caption{Adaptive authority allocation weight $\beta^t$ versus inference confidence $b$ and distance to goal $d$. Warmer/cooler colors indicate higher user/robot authority. $b_{\text{thresh}} = 0.6$, $d_{\min} = 0.1 \times d_{\text{workspace}}$, $d_{\max} = 0.8 \times d_{\text{workspace}}$.}
    \label{fig:beta_heatmap}
\end{figure}

The combined effect of these modulations is visualized in Figure~\ref{fig:beta_heatmap}, which shows how $\beta^t$ varies across the confidence-distance state space.

Together, these two modulations implement an intuitive policy. The robot takes assertive legible control only when it is both confident in its inference and close to the goal, while deferring to the user otherwise. This ensures legible communication is strongest precisely when it is most informative, near the goal where goal discrimination matters most.

\section{SIMULATION STUDY}
\label{sec:experiments}

We validate legible shared autonomy through a 2D simulation user study with 20 participants, testing the hypothesis that legibility weight affects transparency (understanding and prediction accuracy), subjective experience (intuitiveness and collaboration quality), and control effort.

\subsection{Experimental Design} 
\label{sec:sim_study}

\subsubsection{Task and Environment}

\begin{figure}[h]
    \centering
    \includegraphics[width=1\linewidth]{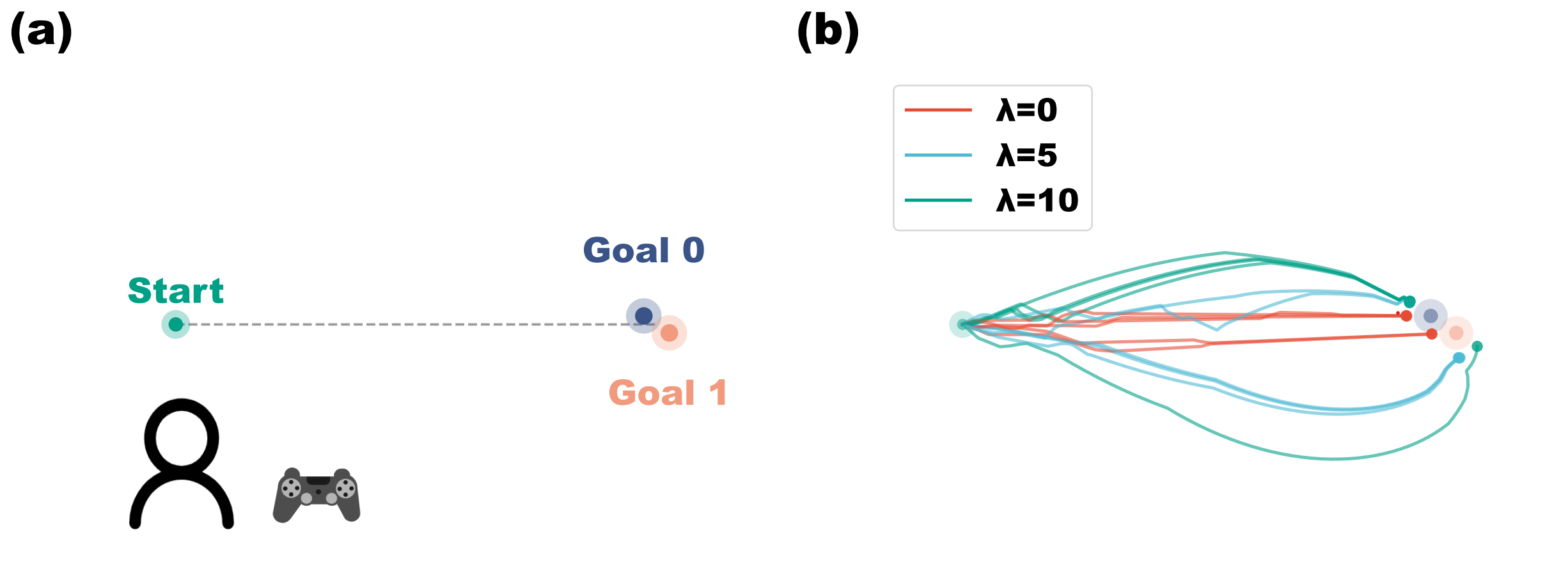}
    \caption{Experimental setup showing (a) workspace configuration with start position and two closely spaced goal regions, and (b) example trajectories from one participant across all conditions.}
    \label{fig:experimental_setup}
\end{figure}

Participants controlled a circular agent in a 2D workspace (800$\times$600 pixels) using a game controller's left analog stick, as illustrated in Figure~\ref{fig:experimental_setup}. The workspace contained two closely spaced objects on the right side: Goal 0 at [650, 290] and Goal 1 at [680, 310], separated by only 30 pixels horizontally and 20 pixels vertically. From the start position [100, 300] on the left side, both objects appear in nearly identical directions, causing efficient assistive motion toward either object to remain directionally ambiguous. Both user and robot actions were velocity-limited to 50 pixels per second, and the system operated at 30 Hz. Workspace size was defined as the diagonal $d_{\text{workspace}} = \sqrt{800^2 + 600^2} = 1000$.

\subsubsection{Conditions}

We tested three legibility weight conditions ($\lambda$ in Equation~\ref{eq:optimization}): Standard SA ($\lambda = 0$), Medium Legibility ($\lambda = 5$), and High Legibility ($\lambda = 10$). All conditions used identical adaptive authority allocation parameters ($\beta_{\text{base}} = 0.6$, $\beta_{\text{min}} = 0.2$, $b_{\text{thresh}} = 0.6$, $d_{\text{min}} = 0.1 \times d_{\text{workspace}}$, $d_{\text{max}} = 0.8 \times d_{\text{workspace}}$).

\subsubsection{Procedure}

Participants completed two phases without knowledge of legibility parameters. 
\textbf{Phase 1:} 15 trials (5 per condition, randomized) where the system paused at 50\% progress for transparency questions. \textbf{Phase 2:} 3 trials (1 per condition, sequential) followed by subjective ratings.
 
\subsubsection{Dependent Measures}
\label{sec:measures}

\textbf{Transparency} (Phase 1). At each trial midpoint, participants answered two questions. First, ``Did the robot understand your intention?'' (Yes/No). Second, ``What goal did the robot infer?'' (Goal 0/Goal 1). Responses to the first question yielded the understanding rate, and correct responses to the second question were scored as prediction accuracy.

\textbf{Subjective experience} (Phase 2). After each condition trial, participants rated the system on two 10-point scales (1 = lowest, 10 = highest) for intuitiveness (``How intuitive was the robot's behavior?'') and collaboration quality (``How well did you collaborate with the robot?'').

\textbf{Control effort} (Phase 1). We measured average user input magnitude per trial:
\begin{equation}
\text{Control Effort} = \frac{1}{n} \sum_{t=1}^{n} \|\mathbf{a}_H^t\|_2
\label{eq:control_effort}
\end{equation}
where $\mathbf{a}_H^t$ is the user input at timestep $t$ and $n$ is the number of timesteps. Higher values indicate greater physical workload.

\subsection{Results}
\label{sec:results}

\begin{figure*}[t]
    \centering
    \includegraphics[width=1\textwidth]{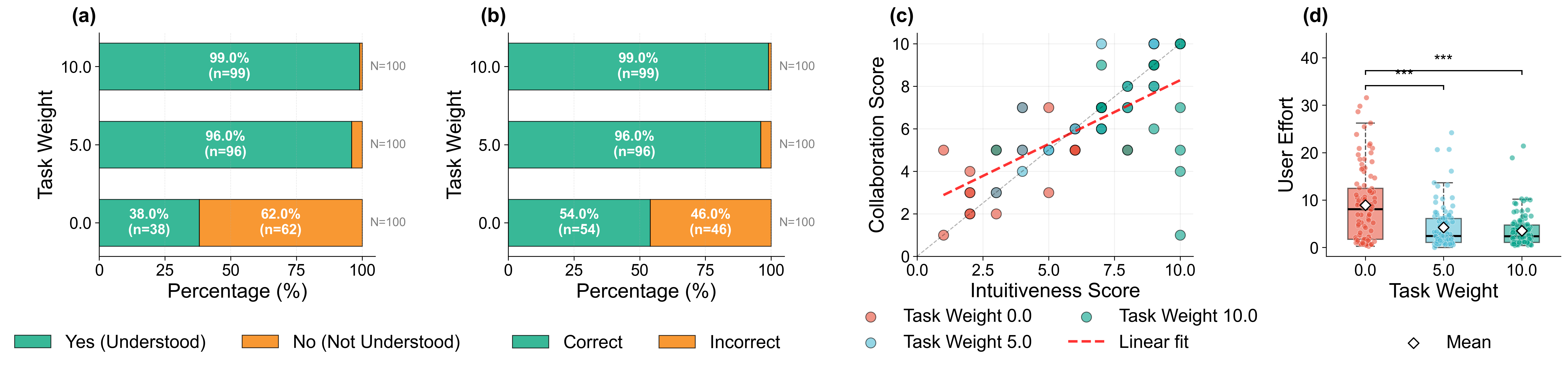}
   \caption{2D simulation study results across three legibility conditions ($\lambda \in \{0, 5, 10\}$). (a) Understanding rates. (b) Prediction accuracy. (c) Subjective ratings correlation. (d) Control effort. See Section~\ref{sec:results} for statistical analysis. $^*p < 0.05$, $^{**}p < 0.01$, $^{***}p < 0.001$.}
    \label{fig:compact_results}
\end{figure*}

Panels (a), (b), and (d) of Figure~\ref{fig:compact_results} present Phase 1 results from 300 trials (20 participants $\times$ 3 conditions $\times$ 5 trials). Panel (c) presents Phase 2 intuitiveness and collaboration ratings from 60 observations (20 participants $\times$ 3 conditions). For statistical analysis, participants were treated as repeated-measures units. Phase 1 trial-level measures, including understanding responses, prediction responses, and control effort, were averaged across the five trials completed by each participant in each condition; binary responses were therefore analyzed as per-participant rates. Phase 2 intuitiveness and collaboration ratings required no preprocessing, with one score recorded per participant under each condition. Friedman tests were applied across the three conditions; when the overall effect was significant, pairwise Wilcoxon signed-rank tests with Bonferroni correction were conducted for post hoc comparisons.

\subsubsection{Transparency Metrics}

Understanding rate differed significantly across conditions ($\chi^2(2) = 28.10$, $p < 0.001$). In Standard SA, only 38.0\% (38/100 trials) of participants could perceive that the robot understood their intention at mid-trial, compared to 96.0\% (96/100) in Medium Legibility and 99.0\% (99/100) in High Legibility. Pairwise comparisons showed Standard SA differed significantly from both Medium ($p_{\mathrm{corr}} < 0.001$) and High Legibility ($p_{\mathrm{corr}} < 0.001$), while Medium and High did not differ ($p_{\mathrm{corr}} = 0.250$).

Prediction accuracy showed a similar pattern ($\chi^2(2) = 29.61$, $p < 0.001$). Participants correctly identified the robot's inferred goal in 54.0\% (54/100) of Standard SA trials versus 96.0\% (96/100) in Medium Legibility and 99.0\% (99/100) in High Legibility. The same pairwise pattern emerged, where Standard SA differed significantly from both Medium ($p_{\mathrm{corr}} = 0.001$) and High Legibility ($p_{\mathrm{corr}} < 0.001$), while Medium and High did not differ ($p_{\mathrm{corr}} = 0.307$).

These results demonstrate that legible motion can effectively convey the robot's internal inference state to users. Compared with Standard SA, the legible motion design significantly improves users' ability to perceive whether the robot understands their intention and to correctly identify the inferred goal.

\subsubsection{Subjective Ratings}

Intuitiveness ratings differed significantly across conditions ($\chi^2(2) = 28.97$, $p < 0.001$). Mean ratings were $3.90 \pm 2.10$ (Standard SA), $6.70 \pm 2.03$ (Medium Legibility), and $8.35 \pm 1.73$ (High Legibility).

Collaboration ratings showed a similarly strong effect ($\chi^2(2) = 17.94$, $p < 0.001$). Mean ratings were $4.25 \pm 1.80$ (Standard SA), $7.00 \pm 2.15$ (Medium Legibility), and $6.95 \pm 2.24$ (High Legibility). Notably, Medium and High Legibility yielded comparable collaboration scores, suggesting that moderate legibility is sufficient to produce a qualitative shift in perceived collaboration quality.

Intuitiveness and collaboration ratings correlated positively (Pearson $r = 0.66$, $p < 0.001$; Spearman $\rho = 0.66$, $p < 0.001$), as shown in Figure~\ref{fig:compact_results}(c). However, we observed an interesting yet reasonable phenomenon: several data points in the lower-right region of Figure~\ref{fig:compact_results}(c) indicate that while $\lambda = 10$ produced highly intuitive motion, some participants reported lower collaboration scores for this condition. Qualitative feedback revealed that these participants found the robot's behavior overly assertive, preferring the more balanced assistance of $\lambda = 5$ despite its slightly lower clarity. This suggests that making robot intent as explicit as possible does not necessarily translate into a better collaborative experience, and that an optimal shared autonomy system should balance communicative clarity with behavioral subtlety, rather than maximizing legibility alone.

\subsubsection{Control Effort}

Control effort differed significantly across conditions ($\chi^2(2) = 21.70$, $p < 0.001$). Pairwise Wilcoxon signed-rank tests with Bonferroni correction confirmed that both Medium Legibility ($p_{\mathrm{corr}} < 0.001$) and High Legibility ($p_{\mathrm{corr}} < 0.001$) produced significantly lower user input magnitude than Standard SA, while no significant difference was found between the two legible conditions ($p_{\mathrm{corr}} = 0.342$). This saturation pattern mirrors the transparency metrics, suggesting that moderate legibility achieves most of the workload reduction benefit.

\section{PHYSICAL ROBOT STUDY}
\label{sec:robot_study}

We validate legible shared autonomy on a physical robot through a user study with 15 participants using a six-degree-of-freedom robot arm, testing whether legible motion improves subjective experience (intuitiveness and collaboration quality) and reduces control effort while maintaining task efficiency.

\subsection{Experimental Design}
\label{sec:robot_design}

\subsubsection{Task and Environment}

\begin{figure}[h]
    \centering
    \begin{subfigure}[b]{0.5\linewidth}
        \centering
        \includegraphics[width=\linewidth]{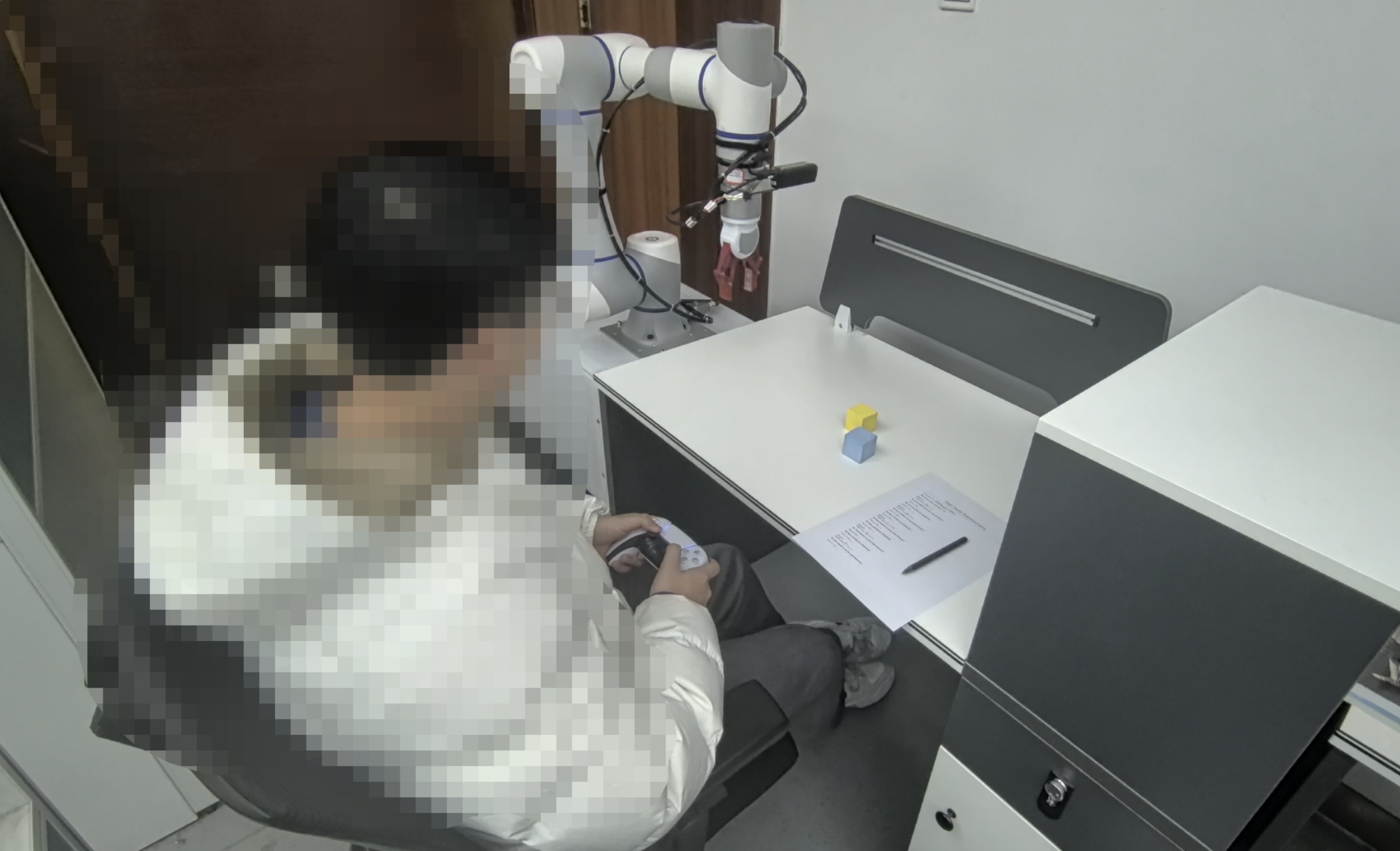}
        \caption{Physical robot setup}
        \label{fig:robot_env}
    \end{subfigure}
    \hfill
    \begin{subfigure}[b]{0.45\linewidth}
        \centering
        \includegraphics[width=\linewidth]{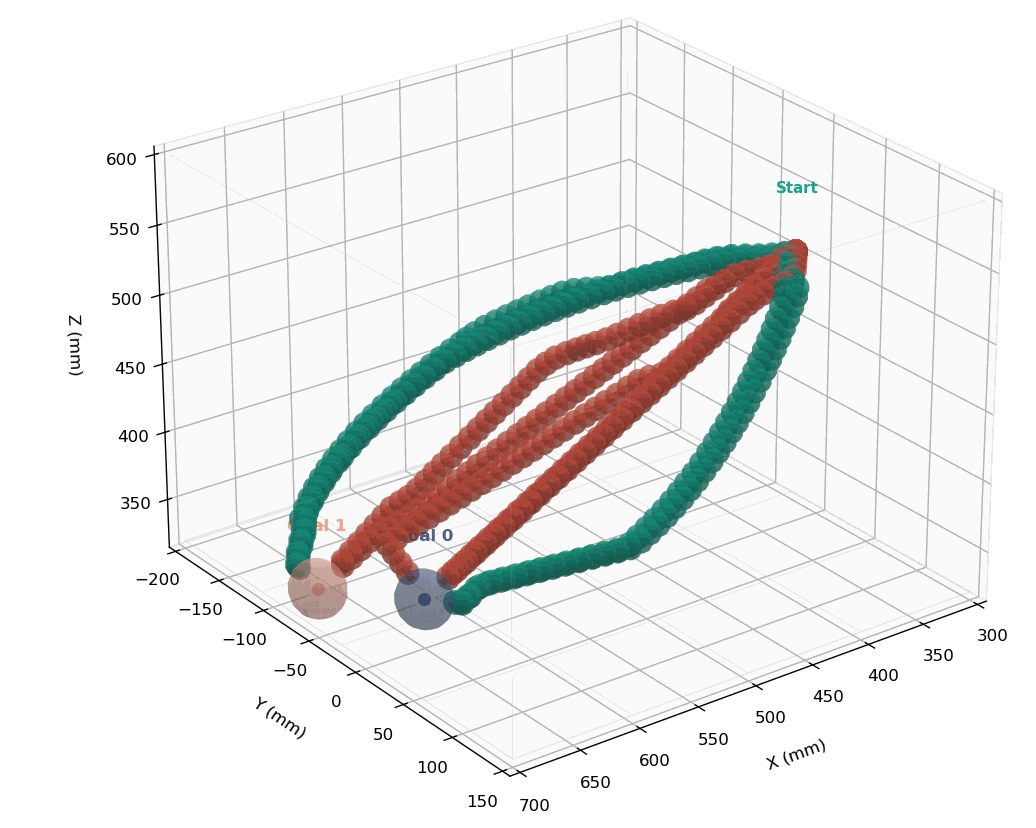}
        \caption{Example trajectories}
        \label{fig:robot_traj}
    \end{subfigure}
    \caption{Physical experimental setup showing (a) The Dobot CR5 collaborative robot arm and workspace configuration showing two closely spaced target objects on the table. (b) Representative trajectories demonstrating Standard SA and Legible SA motion patterns during reaching tasks.}
    \label{fig:robot_setup}
\end{figure}

Participants controlled a Dobot CR5 collaborative robot arm using a game controller to reach one of two closely spaced objects (small boxes, 8 cm apart) on a table, as shown in Figure~\ref{fig:robot_setup}. The objects were positioned such that initial approach directions from the home configuration remained nearly identical, replicating the directional ambiguity of the simulation study. Example trajectories for both conditions are illustrated in Figure~\ref{fig:robot_traj}. The robot operated at 30~Hz with velocity limits of 5~cm/s in a workspace of approximately 60~cm $\times$ 40~cm $\times$ 30~cm.

\subsubsection{Conditions}

We compared Standard SA ($\lambda = 0$) and Legible SA ($\lambda = 1$). All other system parameters remained identical to Section~\ref{sec:sim_study}.

\subsubsection{Procedure}

Participants completed 10 trials (5 per condition, randomized order, randomly assigned goals) without knowledge of condition manipulations, providing subjective ratings after each trial. Unlike the simulation study, we did not interrupt the physical robot trials with midpoint transparency questions because stopping the manipulator during execution would disrupt continuous interaction and raise safety risks for participants.

\subsubsection{Dependent Measures}
Following each trial, participants evaluated the system using two 10-point Likert scales (where 1 represents the lowest score and 10 the highest). The evaluation assessed intuitiveness through the question ``How intuitive was the robot's behavior?'' and collaboration quality through ``How well did you collaborate with the robot?''

\subsection{Results}
\label{sec:robot_results}

\begin{figure*}[t]
    \centering
    \includegraphics[width=\textwidth]{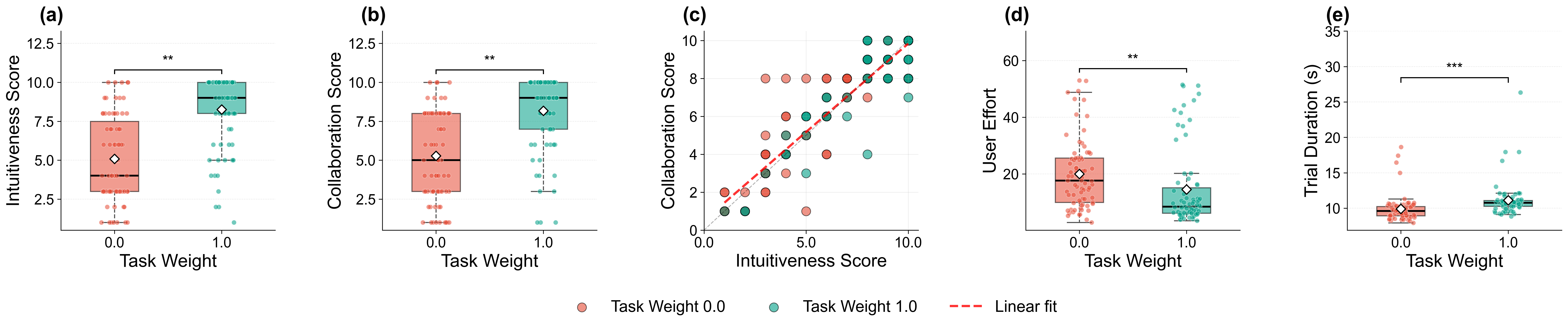}
    \caption{Physical robot study results across two conditions ($\lambda \in \{0, 1\}$). (a) Intuitiveness scores. (b) Collaboration scores. (c) Subjective ratings correlation. (d) Control effort. (e) Task duration. See Section~\ref{sec:robot_results} for statistical analysis. $^*p < 0.05$, $^{**}p < 0.01$, $^{***}p < 0.001$.}
    \label{fig:robot_results}
\end{figure*}

Panels (a), (b), (d), and (e) of Figure~\ref{fig:robot_results} present objective and subjective results from 150 trials (15 participants $\times$ 2 conditions $\times$ 5 trials). Panel (c) presents the correlation between subjective ratings. For statistical analysis of panels (a)(b)(d)(e), participants were treated as repeated-measures units. Trial-level intuitiveness ratings, collaboration ratings, control effort, and task duration were averaged across the five trials completed by each participant in each condition. Wilcoxon signed-rank tests were then applied across the two conditions, while Pearson/Spearman correlations were computed for the rating relationship in panel (c).

\subsubsection{Subjective Ratings}

Figure~\ref{fig:robot_results}(a) and (b) reveal the transformative impact of legible motion on user experience. Intuitiveness ratings (Figure~\ref{fig:robot_results}(a)) increased dramatically from $5.08 \pm 1.48$ in Standard SA to $8.25 \pm 2.06$ in Legible SA, representing a 62\% improvement ($W = 5.50$, $p = 0.003$). This substantial gain directly validates our core innovation: by optimizing robot actions for both task performance and discriminability (Eq.~\ref{eq:optimization}), the robot's motion becomes interpretable, allowing users to perceive which goal the robot has inferred.

The standard deviations indicate substantial variability across participants, which is expected in physical interaction because users differ in control style and sensitivity to motion cues. Despite this variability, the paired improvement in intuitiveness remained significant, supporting that the action-level legibility metric (Eq.~\ref{eq:legibility}) made robot beliefs easier to perceive across diverse users.

Collaboration quality ratings (Figure~\ref{fig:robot_results}(b)) show parallel improvement, increasing from $5.28 \pm 1.49$ to $8.17 \pm 1.92$, a 55\% gain ($W = 5.00$, $p = 0.003$). This convergence between intuitiveness and collaboration is theoretically significant: it confirms that transparency is not merely an auxiliary feature but a foundational requirement for effective human-robot teaming.

Figure~\ref{fig:robot_results}(c) quantifies the fundamental coupling between perceived understanding and collaboration quality through a striking positive correlation (Pearson $r = 0.926$, Spearman $\rho = 0.910$, both $p < 0.001$). This exceptionally strong relationship ($r^2 = 0.857$) indicates that 86\% of the variance in collaboration scores can be explained by intuitiveness alone, establishing perceived robot understanding as the primary driver of collaborative experience.

Standard SA data points (red) cluster in the lower-left region, showing both low intuitiveness and poor collaboration simultaneously. This provides evidence that efficient but ambiguous assistance fails on both dimensions. Legible SA points (teal) concentrate in the upper-right, demonstrating that motion-based transparency elevates both understanding and partnership quality.

\subsubsection{Control Effort}

Figure~\ref{fig:robot_results}(d) provides crucial objective validation that perceived understanding translates into behavioral change. Control effort decreased significantly from $19.96 \pm 10.39$ to $14.50 \pm 13.41$, a 27\% reduction ($W = 12.00$, $p = 0.004$). This result directly confirms our hypothesis that legible motion enables users to recognize correct inference and delegate control earlier, reducing unnecessary intervention.

In Standard SA, users cannot make this determination because ambiguous motion provides no discriminative information, forcing them to maintain control throughout the trial as a conservative safety strategy.

The 27\% effort reduction, while more modest than the simulation study's 47\%, remains highly significant in physical interaction where control fatigue directly affects user comfort and long-term system adoption.

\subsubsection{Task Duration}

Figure~\ref{fig:robot_results}(e) reveals the inherent trade-off in legible motion: task duration increased from $9.91 \pm 1.12$~s to $11.13 \pm 1.46$~s, a 12\% increase ($W = 0.00$, $p < 0.001$). This temporal cost arises because trajectories prioritizing discriminability (high $\mathcal{L}$ in Eq.~\ref{eq:optimization}) deviate from shortest paths, as illustrated in Figure~\ref{fig:robot_setup}(b). When $\lambda > 0$, the robot may initially move slightly away from the direct path to establish clear goal identity before converging.

However, this duration increase must be interpreted within the broader performance profile. The 12\% temporal cost is substantially offset by the 27\% reduction in control effort and 62\% improvement in perceived intuitiveness. For assistive robotics targeting users with motor impairments, physical workload reduction and system comprehensibility often matter more than raw completion speed. Users prefer a system that completes tasks in 11 seconds with minimal effort and high confidence over one that finishes in 10 seconds but requires constant vigilant control.

These results validate the core thesis that robots can effectively communicate their inferred beliefs through motion characteristics in physical interaction. The convergence of subjective experience (how participants felt) and objective behavior (what participants did) provides strong evidence that legible motion transformed understanding into action. Participants both perceived robot understanding and modified their control strategy accordingly, achieving the bidirectional transparency that distinguishes our approach from traditional shared autonomy. 

\section{CONCLUSION}
This work introduced legible shared autonomy, a shared-control framework in which assistive motion helps complete the task while revealing the robot's inferred user goal. The method combines an action-level legibility metric with confidence-aware authority allocation, enabling the robot to communicate its belief through motion while preserving user control when inference confidence is low.

The simulation and physical robot studies show that legible motion can make shared autonomy more interpretable. In simulation, legible assistance improved users' ability to judge whether the robot understood their intent and which goal it had inferred. On the physical robot, legible assistance improved perceived intuitiveness and collaboration quality and reduced control effort, although it increased task duration. These results suggest that legibility is useful for transparent assistance, but should be tuned to the task rather than maximized indiscriminately.

This work also has limitations. We did not directly evaluate high-confidence incorrect inference; in such cases, legible and high-authority assistance could make an incorrect robot belief more salient and harder for the user to ignore. In addition, our experiments used two closely spaced targets, so generalization to cluttered scenes, many candidate goals, or targets arranged along the same approach direction remains open. Despite these limitations, the central finding remains that assistive motion can carry useful information about the robot's inferred belief state, helping shared autonomy move from silent assistance toward more transparent human-robot collaboration\cite{che2020efficient, hemmer2025complementarity}.

\bibliographystyle{IEEEtran}
\bibliography{IEEEabrv,citation}

\end{document}